\documentclass[journal]{IEEEtran} 

\usepackage{amsmath,amssymb,amsfonts,bm,mathdots,amsthm}

\usepackage{comment}

\usepackage{graphicx}
\usepackage{epsfig}
\usepackage{epstopdf}
\usepackage{float}
\graphicspath{{Pictures/}{../jpeg/}} 

\usepackage{array}
\usepackage{multirow}
\usepackage{booktabs}
\usepackage{graphicx}
\usepackage{longtable}
\usepackage{adjustbox}
\usepackage{booktabs}
\usepackage{multirow}
\usepackage{tabularx}
\usepackage{pifont}
\newcommand{\cmark}{\ding{51}}
\newcommand{\xmark}{\ding{55}}

\usepackage{algorithm}
\usepackage{algpseudocode} 

\usepackage[font=small,skip=1pt]{caption}
\ifCLASSOPTIONcompsoc
  \usepackage[caption=false,font=normalsize,labelfont=sf,textfont=sf]{subfig}
\else
  \usepackage[caption=false,font=footnotesize]{subfig}
\fi

\usepackage{xcolor,soul}
\usepackage[table]{xcolor}
\definecolor{rowgray}{gray}{0.93}
\definecolor{rowblue}{RGB}{204,229,255}
\definecolor{lightblue}{RGB}{0,120,200}
\definecolor{darkgreen}{RGB}{0,100,0}
\definecolor{light}{rgb}{0.5, 0.5, 0.5}
\definecolor{deepblue}{RGB}{0,70,140}
\definecolor{deepblue}{RGB}{0,0,139}

\usepackage{cite}
\usepackage{hyperref}
\hypersetup{hidelinks=true}

\usepackage{gensymb}
\usepackage{textcomp}
\usepackage{pifont}
\usepackage{enumitem}
\usepackage{varioref}
\usepackage{latexsym}
\usepackage[active]{srcltx} 
\usepackage{amssymb}

\def\BibTeX{{\rm B\kern-.05em{\sc i\kern-.025em b}\kern-.08em
    T\kern-.1667em\lower.7ex\hbox{E}\kern-.125emX}}

\begin{document}
\title{Multi-Head Attention based interaction-aware architecture for Bangla Handwritten Character Recognition: Introducing a Primary Dataset}

\author{
Mirza Raquib$^{*,1}$,
Asif Pervez Polok$^{*,2}$,
Kedar Nath Biswas$^{3}$,
Farida Siddiqi Prity$^{4}$,
Saydul Akbar Murad$^{5}$,
Nick Rahimi$^{**,5}$%
\thanks{$^{*}$These authors contributed equally to this work.}%
\thanks{$^{**}$Corresponding author.}%
\thanks{$^{1}$Department of Computer and Communication Engineering,
International Islamic University Chittagong, Bangladesh.
Email: mirzaraquib@iiuc.ac.bd}%
\thanks{$^{2}$mPower Social Enterprise.
Email: asifpolok.research@gmail.com}%
\thanks{$^{3}$Department of Information and Communication Engineering,
Noakhali Science and Technology University, Bangladesh.
Email: kedarnathbiswas06@gmail.com}%
\thanks{$^{4}$Department of CSE, Netrokona University, Netrokona, Bangladesh. Email: prity@neu.ac.bd}%
\thanks{$^{5}$School of Computing Sciences and Computer Engineering, University of Southern Mississippi, USA.
Email: saydulakbar.murad@usm.edu, nick.rahimi@usm.edu}%
}

\maketitle

\begin{abstract}
Character recognition is the fundamental part of an optical character recognition (OCR) system. Word recognition, sentence transcription, document digitization, and language processing are some of the higher-order activities that can be done accurately through character recognition. Nonetheless, recognizing handwritten Bangla characters is not an easy task because they are written in different styles with inconsistent stroke patterns and a high degree of visual character resemblance. The datasets available are usually limited in intra-class and inequitable in class distribution. A significant number of datasets do not provide proper information about the data collection method and the demographics of the contributors. These constraints have complicated the formulation of strong models that can be applied in reality. We have constructed a new balanced dataset of Bangla written characters to overcome those problems. This consists of 78 classes and each class has approximately 650 samples. It contains the basic characters, composite (Juktobarno) characters and numerals. The samples were a diverse group comprising a large age range and socioeconomic groups. Elementary and high school students, university students, and professionals are the contributing factors. The sample also has right and left-handed writers. We also proposed an interaction-aware, hybrid, multi-head attention deep learning architecture for recognition. The architecture combines EfficientNetB3, Vision Transformer, and Conformer modules, all executed in parallel. The feature representations interact with each other through a multi-head cross-attention fusion module. The suggested model shows strong scores across various assessment indicators. It not only has a high level of generalization, achieving 98.84 accuracy on the proposed dataset and 96.49 accuracy on the external benchmark CHBCR dataset. The Grad-CAM images show which parts of the image contribute to the predictions. The proposed data structure and hybrid architecture can both provide a viable solution for Bangla handwritten character recognition and help build a useful Bangla OCR system. The dataset and source code of this research is publicly available at: {\color{lightblue}\url{https://huggingface.co/MIRZARAQUIB/Bangla_Handwritten_Character_Recognition}}.

\end{abstract} 

\section{Introduction}
\label{sec:1}
The increasing demand for large-scale document digitization has made optical character recognition (OCR) an essential component of modern document analysis systems. OCR converts scanned images of characters into machine-readable representations and supports various real-world applications such as archival preservation, financial documentation, postal services, and automated examination systems \cite{memon2020handwritten}. Handwritten Character Recognition (HCR) is one of the sub-areas of OCR. Recognition of typeset characters is inherently less challenging than decoding handwritten text, due to large inter-individual disparities. The size, stroke breaks, curves, and the spatial arrangement of characters are highly heterogeneous among writers. These factors increase the difficulty of handwritten character recognition \cite{bappi2024bnvglenet}. Bangla handwritten character recognition (BHCR) introduces additional structural challenges. The Bangla script contains vowels, consonants, numerals, and compound characters formed by combining consonants. Many characters differ only by a small stroke or matra, which increases inter-class similarity and reduces class separability \cite{chowdhury2020bangla}. Writer-dependent factors such as age, educational background, and writing habits also introduce significant variation in handwriting styles \cite{rahman2019bangla}. These variations create complex and diverse stroke patterns within the same character class. Consequently, BHCR becomes a fine-grained classification problem, which demands powerful feature representation and discrimination.

The development of handwritten character recognition in common languages such as English has advanced significantly. There are several effective OCR programs that exist to perform this task. On the contrary, BHCR is considerably less developed. This gap limits the effectiveness of automated processing for Bangla documents compared to other major languages. Character recognition enables higher-level tasks such as word recognition, sentence transcription, and document digitization. Errors at the character level often propagate to later stages and reduce the overall performance of OCR systems. Thus, proper character recognition is key to improving the performance of the Bangla OCR applications.  High quality dataset remains a significant challenge for BHCR research. Nevertheless, most Bangla datasets lack variety in writing styles and a balanced representation of classes. The data collection procedure is inadequately documented in several instances, and details regarding the writing conditions or sources are often lacking. Certain datasets are gathered by a few writers or by a small age range. This does not reflect the variation of handwriting in the real world. Age, writing habits, stroke thickness, the location of matras, and personal writing styles are hence not sufficiently represented. Bangla is also regarded as a low-resource language, with large and diverse datasets of handwritten text, yet resources remain scarce. Therefore, a balanced, rich dataset, along with an effective recognition system, should be developed to build stable models and enhance downstream Bengali OCR systems.

Previous works shifted handwritten character recognition from manually designed features to convolutional neural network (CNN) models. The CNN architecture also proved very powerful. It automatically extracts local spatial features from images\cite{ahlawat2020improved,kavitha2022benchmarking,mushtaq2021urdudeepnet}. Later, deeper pretrained architectures such as ResNet and Inception further improved representation learning for offline handwriting recognition \cite{Chandankhede2023Offline,ghosh2021performance}. Lightweight models such as ShuffleNet and SqueezeNet were introduced to reduce computational complexity while maintaining competitive performance \cite{el2025leveraging}. Nonetheless, CNN models primarily focus on local receptive fields and are often unable to represent long-range structural links between distant parts of characters. Recently, transformer-based methods have been proposed to reduce this limitation. Vision transformers and graph-based transformer models capture global dependencies among character units. They provide better character representations and more reliable contextual information \cite{li2025htr,gan2023characters}. Attention mechanisms, e.g., self-attention networks and vertical attention models, also enhance the contextual perception of handwritten text further \cite{molavi2024self,coquenet2022end}. CNN and transformer hybrid architectures have also been studied to combine local and global feature representations. Attention-based transfer learning has also been shown to achieve higher efficiency in handwriting recognition than before \cite{fateh2024advancing}. This directly impacts the performance of downstream OCR tasks such as word recognition and document analysis.

Despite these improvements, the present research continues to face two severe problems. First, the existing datasets often fail to capture real-world handwriting variability. These limits the ability to capture natural handwriting. Second, most hybrid recognition models rely on either simple feature concatenation or ensemble methods, which do not explicitly capture how heterogeneous feature representations interact. This limitation becomes more significant for Bangla characters, where small stroke differences and compound structures require both local and global understanding. A high-quality dataset is essential for developing reliable recognition models and supporting downstream Bangla OCR applications. An interaction-sensitive robust architecture capable of maintaining localized stroke representations and global structural constraints is also needed. 

The paper attempts to address such problems by proposing a balanced dataset of Bangla handwritten characters and a hybrid deep neural network recognition architecture. The dataset has significant intra-class variation and contributors from various age groups and social backgrounds, including school students, university students, and professionals. Such a strategy represents a broad portfolio of handwriting, as well as natural variations of writing. This diversity helps capture natural handwriting variations. The proposed parallel hybrid architecture combines convolutional networks for local stroke features and transformer-based modules for global structural relationships. The limitation of feature concatenation is overcome by having a multi-head cross-attention fusion mechanism to allow interaction between feature streams. It is a design that is interaction-sensitive, enhancing the recognition of structurally similar and compound characters. The applications supported by the framework include document digitization, archiving of Bangla documents, automated inspection evaluation, and a language processing system. Enormous trials involve comparison of the backbone, preprocessing analysis, per-class analysis, five-fold cross-validation and external dataset testing. The proposed framework improves recognition accuracy while maintaining strong generalization and robustness across diverse handwriting styles.

\textbf{Major Contributions of This Work}

\begin{itemize}

\item We construct a balanced Bangla handwritten character dataset with 78 classes and significant diversity across writers, including different age groups, genders, left-handed, and right-handed users.

\item We propose a parallel hybrid architecture integrating EfficientNetB3, Vision Transformer, and Conformer branches for character recognition.

\item We introduce a multi-head cross-attention fusion mechanism to enable interaction between heterogeneous feature representations.

\item We jointly model local stroke-level patterns and global structural dependencies within a unified framework.

\item We conduct extensive experiments including backbone analysis, preprocessing studies, feature extraction studies, five-fold cross-validation, and evaluation on the external CHBCR\cite{towhid2020chbcrdb} dataset.

\end{itemize}

The rest of this article is organized as follows. Section \ref{sec:2} summarizes the related work. Section \ref{sec:3} presents the proposed methodology, including dataset preparation, preprocessing steps, and the hybrid fusion architecture. Section \ref{sec:4} reports the experimental results and performance evaluation of the proposed model. Finally, Section \ref{sec:5} concludes the paper.

\section{Literature Review}
\label{sec:2}
Bangla handwritten character recognition (BHCR) research has become a growing scholarly interest over the past couple of years. It is up to date with advances in digital document processing and language technology. The initial research primarily involved conventional machine learning models and shallow neural network models to identify independent Bangla characters. Subsequently, the most popular ones were convolutional neural networks (CNNs), which are capable of automatically extracting stroke patterns and the structure of handwritten images. Different CNN architectures, as well as pretrained models, have been tested in various works to enhance recognition accuracy. Transformer-based architectures and attention have received recent attention to define global connectivity between character units. Representation learning has similarly been studied using models that condense transformer and convolutional neural network architectures. Nevertheless, the current literature is mainly based on single-stream implementations or rough feature-integration mechanisms. Some studies were tested on datasets with limited diversity of writers or small sample sizes. These drawbacks limit the capacity of these models to trace structural differences and handwriting forms that vary among writers.

\begin{table*}[htbp]
\centering
\renewcommand{\arraystretch}{1.5}
\resizebox{\textwidth}{!}{%
\begin{tabular}{p{1cm}p{2.8cm}cccp{2.8cm}p{4.5cm}p{4.5cm}}
\hline

\multirow{2}{*}{\textbf{Study}} 
& \multirow{2}{*}{\textbf{Dataset Name}}
& \multicolumn{3}{c}{\textbf{Dataset Type}} 
& \multirow{2}{*}{\textbf{Architecture Type}} 
& \multirow{2}{*}{\textbf{Identified Gap}} 
& \multirow{2}{*}{\textbf{Resolution in Proposed Framework}} \\

\cline{3-5}

& & {\textbf{Basic}} 
& {\textbf{Digits}} 
& {\textbf{Compound}} 
&  &  &  \\

\hline

\cite{rabby2018ekushnet}
& Ekush (Primary)
& \cmark & \cmark & \cmark 
& EkushNet (22-Layer DCNN) 
& Fixed CNN architecture without adaptive feature fusion
& Interaction-aware hybrid architecture enabling adaptive feature fusion\\

\cite{islam2022ratnet}
& BanglaLekha-Isolated, ISI, CMATERdb, IUBMCdb
& \cmark & \cmark & \cmark 
& CNN with residual unit and Spatial Attention
& Single-stream convolutional modeling 
& Employes parallel CNN–Transformer streams \\

\cite{chakraborty2024msbnet}
& BanglaLekha, Ekush, CMATERdb
& \cmark & \cmark & \cmark 
& MSBNet (Tri-scale Lightweight CNN)
& No hybrid CNN–Transformer modeling 
& Integrates hybrid CNN–Transformer \\ 

\cite{bappi2024bnvglenet}
& Primary
& \cmark & \cmark & \cmark 
& BNVGLENET (VGG + LeNet) 
& No global self-attention modeling 
& Utilizes multi-head cross-attention modeling \\ 

\cite{opu2024handwritten}
& BanglaLekha, ISI, CMATERdb, Ekush
& \cmark & \cmark & \xmark 
& Lightweight CNN
& Less time consuming architecture but some pretrained models shows better performance
& Improves generalization through balanced primary dataset with diverse writer variability \\ 

\cite{sabira2024bengali}
& BanglaLekha
& \cmark & \cmark & \cmark 
& Morphological preprossing enhanced DCNN+ECA
& Preprocessing dependency 
& Reduces reliance on heavy preprocessing via robust deep feature learning \\ 

\cite{raquib2024vashanet}
& Primary
& \cmark & \xmark & \xmark 
& 26-layer DCNN
& Small dataset and high computational complexity  
& Constructs a balanced large-scale dataset with significant structural variability\\ 

\cite{raquib2024vashanet_v2}
& Primary + CMATERdb
& \cmark & \xmark & \xmark 
& 19-layer lightweight DCNN 
& Single-backbone architecture restricts multi-scale representational diversity
& Enables richer multi-scale representation \\ 

\cite{saha2025inksynth}
& BanglaLekha-Isolated
& \cmark & \cmark & \cmark 
& BengNet (ResNet + Inception)
& No global self-attention modeling 
& Introduces explicit attention-based interaction modeling \\ 

\cite{ahmed2025vision}
& Ekush, Matribhasha
& \cmark & \cmark & \cmark 
& Hybrid-L, Hybrid-S
& Distorted Samples are eliminated by preprocessing
& Handles distortions through dataset diversity and attention-driven feature fusion \\

\cite{rahman2026hqcnn}
& NumtaDB, CMATERdb, Ekush, BanglaLekha-Isolated
& \cmark & \cmark & \cmark 
& Hybrid Quantum-Classical CNN (HQCNN)
& Requires classical simulation of quantum circuits and introduces high computational overhead
& Employs efficient CNN-Transformer attention framework with balanced dataset to achieve scalable and practical feature learning \\
\bottomrule
\end{tabular}%
}

\vspace{0.2cm}
\small
\textbf{Legend:} \cmark = Yes, \xmark = No, ECA = Effective Channel Attention

\caption{Comparative gap analysis and how the proposed framework addresses key architectural and dataset limitations}
\label{tab:literature_rev}

\end{table*}

Deep learning architectures and different benchmark datasets have been employed in research on the recognition of Bangla handwritten characters. An example is the EkushNet study, which presented the Ekush dataset. They proposed a 22-layer DCNN achieving 97.73\% accuracy on the Ekush dataset and 95.01\% on the CMATERdb dataset \cite {rabby2018ekushnet}. Likewise, RATNet is a convolutional network that trains with residual units and spatial attention to learn complex patterns in handwriting. The model was tested on recently gathered modifier and compound character data sets, achieving 97.41\% and 93.42\% accuracy, respectively \cite{islam2022ratnet}. Moreover, multi-scale CNN design has also been tested to discern structural differences in handwritten characters. For example, MSBNet presented a lightweight tri-scale CNN architecture and achieved good results across various public datasets, including Ekush, BanglaLekha, and CMATERdb\cite{chakraborty2024msbnet}. Other studies focus on dataset design and hybrid network structures. BNVGLENet, which combines VGG and LeNet architectures, was trained on a custom dataset containing only basic characters and numerals. It acheives accuracies above 97\% for basic characters and numerics \cite{bappi2024bnvglenet}. Furthermore, lightweight CNN models have been developed for efficient character recognition across various datasets such as ISI, CMATERdb, Ekush, and BanglaLekha. The model reaches around 96.8\% accuracy while maintaining low computational cost \cite{opu2024handwritten}. Another noteworthy approach applies morphological preprocessing with DCNN and channel attention. It improves recognition accuracy to 96.29\% on BanglaLekha isolated characters \cite{sabira2024bengali}. In our earlier work, VashaNet introduced a newly collected dataset of handwritten Bangla basic characters. It only contains basic characters samples collected from 115 students, and employed a 26-layer DCNN architecture. This architecture achieves 94.60\% accuracy \cite{raquib2024vashanet}. A later extension expanded this dataset by adding additional contributors and combining it with CMATERdb samples. This version used a lightweight 19-layer DCNN architecture and achieved 95.20\% accuracy on the merged dataset \cite{raquib2024vashanet_v2}. Most recent studies have also explored hybrid CNN architectures. BengNet, for example, combines ResNet and Inception modules, and achieved 96.66\% accuracy on the Ekush and Matribhasha datasets \cite{saha2025inksynth}. Moreover, hybrid CNN--Transformer frameworks have been investigated recently. HybridNet-S integrates multiple backbones, including EfficientNetBO, DenseNet121, and TinyViT, and achieves about 95.8\% accuracy on the BanglaLekha dataset. \cite{ahmed2025vision}. They also investigate a second HybridNet-L integrates RestNet50, EfficientNetBO and ViT. Recent studies have investigated quantum-enhanced deep learning architectures for Bangla handwritten character recognition. They introduced a Hybrid Quantum-Classical Convolutional Neural Network (HQCNN)\cite{rahman2026hqcnn}, which incorporates random quantum circuits as a quantum convolutional layer for feature extraction. This model was evaluated on multiple public datasets, including NumtaDB, CMATERdb, Ekush, and BanglaLekha-Isolated, achieving up to 99.45\% accuracy on the Ekush numerical dataset and demonstrating superior feature representation compared to classical CNN models.   

Datasets play a significant role in Bangla handwritten character recognition studies. The available datasets have a lot of samples, but usually do not exhibit inter-class diversity and variation in writing styles. In most studies, the way the data are collected is not clearly explained, and details of the writers, writing conditions, and the collecting procedure are usually omitted. Age heterogeneity among contributors is also rarely documented. Most datasets have been collected within a restricted age range or with a limited number of participants. However, there is great variation in Bangla handwriting, which varies with age, writing habits, stroke thickness, the location of matrattes, and each person's own character. Such changes make character discovery a challenge. In the past, it was suggested to use large CNN models, lightweight networks, hybrid architectures, and attention mechanisms. However, most methods still rely on unidirectional stream feature extraction and basic feature fusion. Thus hampering local stroke and disciplined character learning.

We have suggested a new BHCR dataset and a hybrid recognition system to overcome those challenges. It has 50700 samples across 78 classes of Bangla characters. It contains basic characters, compound characters, and numerals. Contributors from various age groups and occupations, including school students and senior individuals, were used to collect handwritten samples. This collection plan encompasses various forms of handwriting and handwriting rituals. A grid-based protocol and high-resolution scanning were used during data acquisition to ensure that fine stroke details were not lost. The dataset captures inter-class variation and is realistic with respect to handwriting variation. A cross-attention deep learning architecture, which is a hybrid one, is presented along with the dataset. The structure is a hybrid of convolutional neural networks and transformer-based models that operate in parallel. Convolutional layers capture local stroke details, and the transformer module framework establishes global structural relations. It is a cross-attention fusion mechanism that allows interaction between feature streams and refines a final representation. Backbone comparison, preprocessing analysis, feature selection study, cross-validation, and evaluation on an external dataset are used to evaluate performance.

\section{Methodology}
\label{sec:3}

\begin{figure*}[t]
\centering
\includegraphics[width=\textwidth]{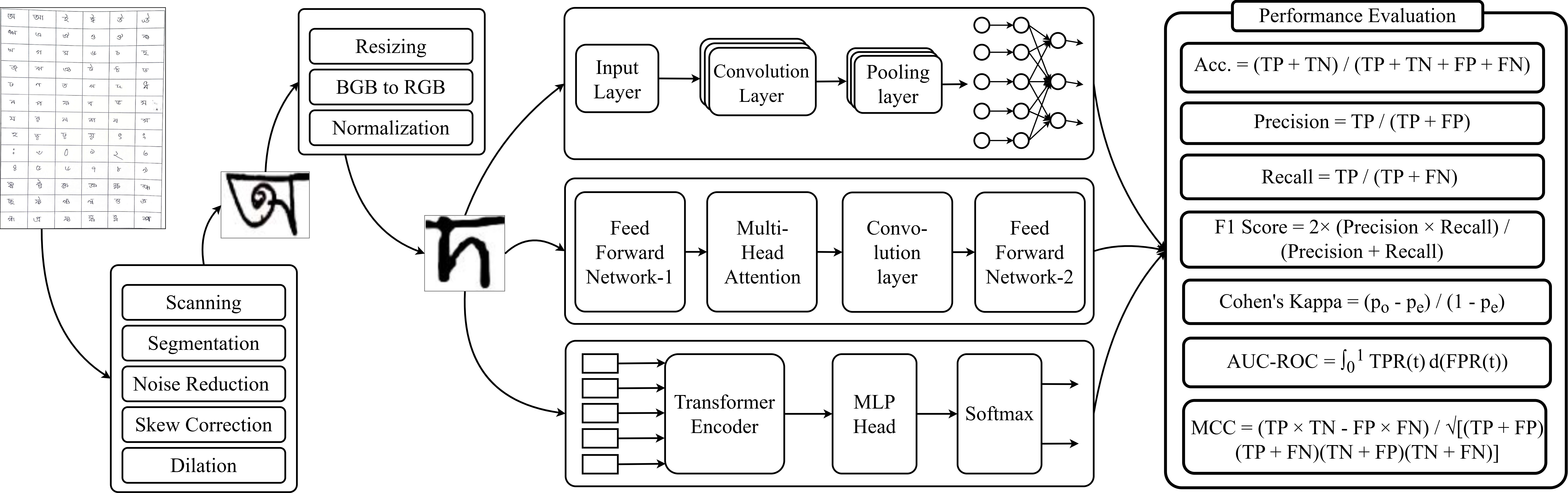}
\caption{Proposed workflow for handwritten character recognition.}
\label{fig-workflow}
\end{figure*}

Recognition of handwritten Bengali characters, including compound characters and numerals, is more problematic.   This similarity has intensified more confusion of classes, hence, making it harder to classify. Thus, enhanced feature learning is now needed. This study has suggested a hybrid deep learning system to recognize Bangla hand written character. The convolutional neural networks and transformer-based attention mechanisms have been integrated in the proposed model. This design enables cross-branch feature interaction and refinement of feature representations. The architecture improves discrimination between structurally similar characters and complex character compositions. We first prepare and standardize the dataset with a structured preprocessing pipeline. Discriminative features are extracted using a CNN branch enhanced by attention modules. In parallel, we model global dependencies using a transformer encoder. We then fuse learned features and classify them into 78 character classes. Figure~\ref{fig-workflow} illustrates the complete workflow of our proposed system.

\subsection{Data Collection and Preparation}
This study extends our previously published Bangla handwritten character datasets\cite{raquib2024vashanet, raquib2024vashanet_v2}. The earlier VashaNet datasets were collected from 150 students. All participants handwrote 50 simple Bangla characters on preformatted paper. This process produced 7,500 handwritten samples. These samples contain only basic Bangla characters. There were no numerals or compound characters in the earlier versions. All samples from the two VashaNet datasets were included in this work as Phase-1 of the data collection process. This study expands the dataset through additional data collection phases. The final dataset contains 50,700 handwritten images across 78 character classes. There are 50 basic Bangla characters, 10 Bangla digits, as well as 18 compound characters. The data collection had four phases plus a minor extension phase. The general data collection procedure of the sample across phases are summarized in Table \ref{tab:datasetsummary}. The distribution of characters by category is also presented in \ref{tab:distributiontable} below. There were 1,700 respondents in the dataset. The contributors are students in grades 6 to 12, university students, and the elderly. The dataset captures handwriting variations across different age groups and educational backgrounds. The age distribution of contributors is presented in Table \ref{tab:age_distribution}. As well, the elaboration of participant demographics was recorded. There are 820 male respondents and 880 female ones. In data collection, writing preference was also considered. There are 1,510 right-handed and 190 left-handed writers. The gender and hand preference statistics are summarized in Table \ref{tab:demographic_distribution}. The participants used a new A4 grid sheet with characters typed in a pre-established grid scheme. One character was placed along every grid cell. This was a structured format that ensured consistent character spacing and made segmenting the document later easier.

\begin{figure*}[t] 
    \centering
    \includegraphics[width=\linewidth]{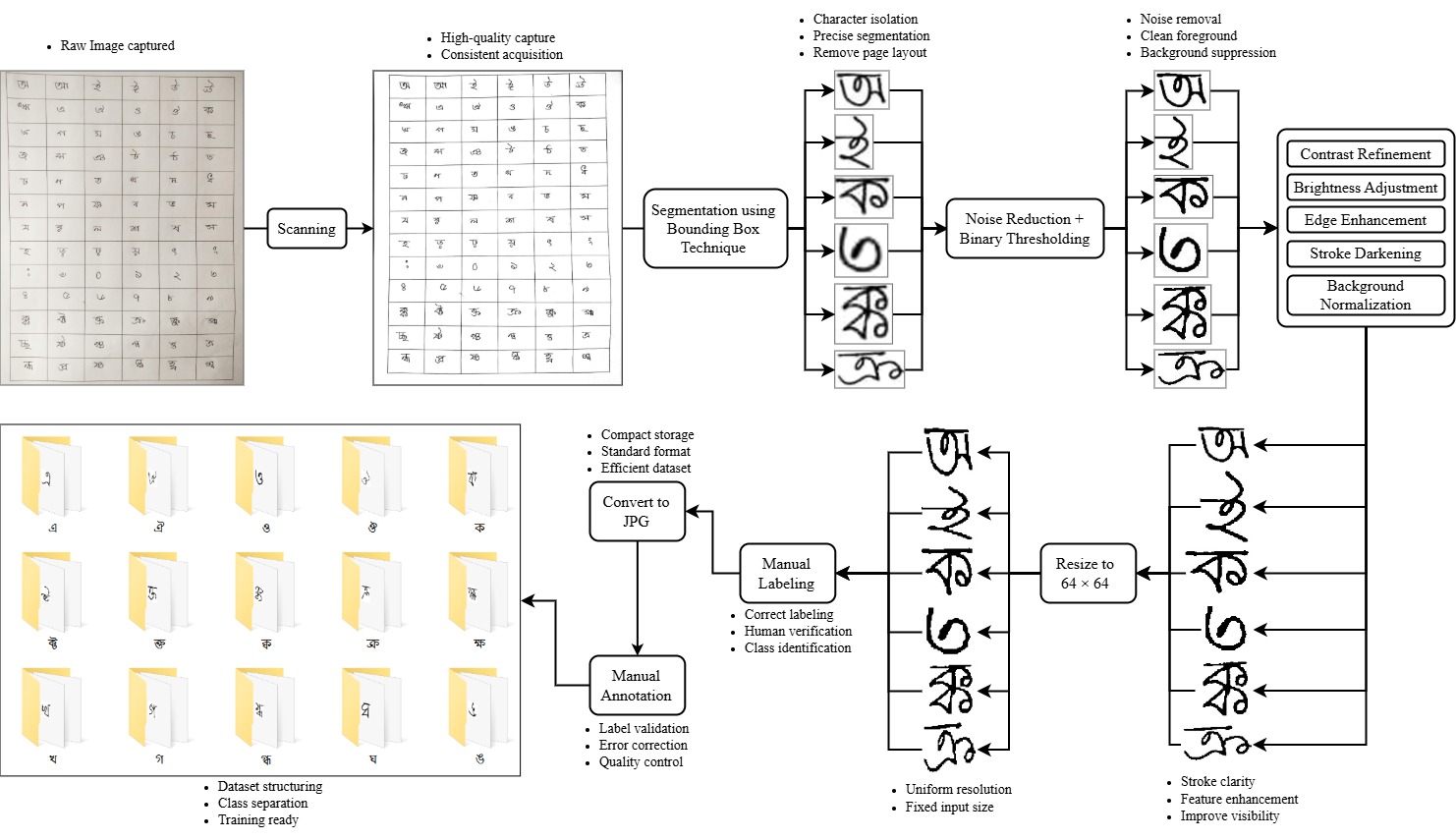}
    \caption{Dataset Preparation Pipeling}
    \label{fig-pipeline}
\end{figure*}

During the data preparation stage, the collected handwritten sheets will be converted into a set of standardized character images. The overall pipeline of dataset preparation is illustrated in Figure~\ref{fig-pipeline}. Each A4 sheet was scanned using a document-scanning application. Characters were written inside grid cells with regular spacing on each sheet. Each grid cell was treated as a character region, and bounding boxes were used to extract individual characters. Background artifacts were removed and processed in noise-reduction and binary-threshold modes, and on the segmented images. There are other refinement operations, such as contrast refinement, brightness adjustment, edge enhancement, stroke darkening, and background normalization, that enhance the visibility of strokes. The pictures were scaled to $64 \times 64$ pixels and changed to JPG. Then the images were marked by hand and authenticated by annotations. Finally, the data set was divided into 78 folders, where each folder contained a character set of Bengali characters. The final dataset contains clean, labeled, and consistently structured character images suitable for Bangla handwritten character recognition experiments.

\begin{table}[h]
\centering
\large
\resizebox{\columnwidth}{!}{
\begin{tabular}{|c|c|c|c|c|}
\hline
\textbf{Phase} & \textbf{Character Type} & \textbf{Participants} & \textbf{Characters Taken} & \textbf{Total Samples} \\ \hline

Phase 1 & Sorobarno+Benjonbarno & 150 & 50 & 7,500 \\ \hline
Phase 2 & Sorobarno+Digits & 500 & 21 & 10,500 \\ \hline
Phase 3 & Benjonbarno & 500 & 39 & 19,500 \\ \hline
Phase 4 & Juktobarno & 650 & 18 & 11,700 \\ \hline
Extension & Digits & 150 & 10 & 1,500 \\ \hline
\rowcolor{rowblue}
\textbf{Total} & \textbf{-} & \textbf{1,700} & \textbf{-} & \textbf{50,700} \\ \hline

\end{tabular}
}
\caption{Summary of Dataset Collection}
\label{tab:datasetsummary}
\end{table}

\begin{table}[h]
\centering
\small
\resizebox{\columnwidth}{!}{
\begin{tabular}{|c|c|c|}
\hline
\textbf{Character Category} & \textbf{Number of Characters} & \textbf{Total Samples} \\ \hline

Sorobarno + Benjonbarno & 50 & 7,500 \\ \hline
Sorobarno & 11 & 5,500 \\ \hline
Digits & 10 & 6,500 \\ \hline
Benjonbarno & 39 & 19,500 \\ \hline
Juktobarno & 18 & 11,700 \\ \hline

\rowcolor{rowblue}
\multicolumn{2}{|c|}{\textbf{Total Samples}} & \textbf{50,700} \\ \hline

\end{tabular}
}
\caption{Distribution of Characters in the Dataset}
\label{tab:distributiontable}
\end{table}

\begin{table}[h]
\centering
\small
\resizebox{\columnwidth}{!}{
\begin{tabular}{|c|c|c|c|}
\hline
\multicolumn{2}{|c|}{\textbf{Gender Distribution}} & \multicolumn{2}{c|}{\textbf{Hand Preference}} \\ \hline
\textbf{Category} & \textbf{Participants (\%)} & \textbf{Category} & \textbf{Participants (\%)} \\ \hline

Male & 820 (48.24\%) & Left Hand & 190 (11.18\%) \\ \hline
Female & 880 (51.76\%) & Right Hand & 1510 (88.82\%) \\ \hline

\rowcolor{rowblue}
\multicolumn{2}{|c|}{\textbf{Total: 1700}} & \multicolumn{2}{c|}{\textbf{Total: 1700}} \\ \hline

\end{tabular}
}
\caption{Gender and Hand Preference Distribution of Participants}
\label{tab:demographic_distribution}
\end{table}

\begin{table}[h]
\centering
\small
\resizebox{\columnwidth}{!}{
\begin{tabular}{|c|c|c|c|}
\hline
\textbf{Age Group} & \textbf{Participants (\%)} & \textbf{Age Group} & \textbf{Participants (\%)} \\ \hline
10--15 & 200 (11.76\%) & 26--30 & 250 (14.71\%) \\ \hline
16--20 & 400 (23.53\%) & 31--35 & 200 (11.76\%) \\ \hline
21--25 & 500 (29.41\%) & 36+ & 150 (8.82\%) \\ \hline
\rowcolor{rowblue}
\multicolumn{4}{|c|}{\textbf{Total Participants: 1700}} \\ \hline
\end{tabular}
}
\caption{Age Group Distribution of Dataset Contributors}
\label{tab:age_distribution}
\end{table}

\subsection{Data Preprocessing}

Data preprocessing plays an important role in handwritten character recognition. Raw scanned images may contain noise, alignment errors, stroke variation, and intensity differences. These issues affect model stability and recognition accuracy. A structured preprocessing pipeline was applied before training. Each step was applied to all 50,700 images.

\textbf{Noise Reduction:}  
Scanned documents often contain salt-and-pepper noise and small pixel variations. A Gaussian filter was applied to reduce noise while preserving structural edges. This filter reduces high-frequency noise while preserving structural edges. Stroke continuity becomes clearer after smoothing. The Gaussian filter \cite{deng1993adaptive} is defined as:

\begin{equation}
G(x,y) = \frac{1}{2\pi\sigma^2} e^{-\frac{x^2 + y^2}{2\sigma^2}}
\end{equation}

Parameter $\sigma$ represents the standard deviation of the Gaussian kernel.

\textbf{Skew Correction:}  
Some scanned pages show small angular misalignment due to manual placement during scanning. Misaligned images affect feature extraction. The Hough transform was used to estimate orientation and correct skew. The Hough transform \cite{xu1993randomized} is defined as

\begin{equation}
\rho = x \cos\theta + y \sin\theta
\end{equation}

The detected angle was used to rotate the image and restore horizontal alignment.

\textbf{Dilation:}  
Handwritten strokes vary in thickness due to writing pressure and style. Thin strokes sometimes become weak after scanning. Morphological dilation was applied to strengthen foreground strokes. This operation enlarges foreground pixels and improves stroke visibility.

\textbf{Resizing:}  
Deep learning models require fixed input dimensions. Character images showed small size differences after segmentation. Each image was resized to a consistent resolution before training. For model training, all images were resized to 128 × 128 pixels.

\textbf{BGR to RGB Conversion:}  
Captured images were stored in BGR color format. Many deep learning frameworks operate with RGB format. Channel ordering was adjusted to convert BGR images into RGB images.

\textbf{Normalization:}  
Pixel intensity values originally ranged from 0 to 255. Pixel normalization scaled the values to the range [0,1] using,

\begin{equation}
X_{norm} = \frac{X}{255}
\end{equation}

Normalized inputs improve training stability and gradient convergence.

The preprocessing pipeline produces aligned and standardized character images. These steps improve feature learning and enhance recognition performance.


\begin{figure*}[t] 
    \centering
    \includegraphics[width=\linewidth]{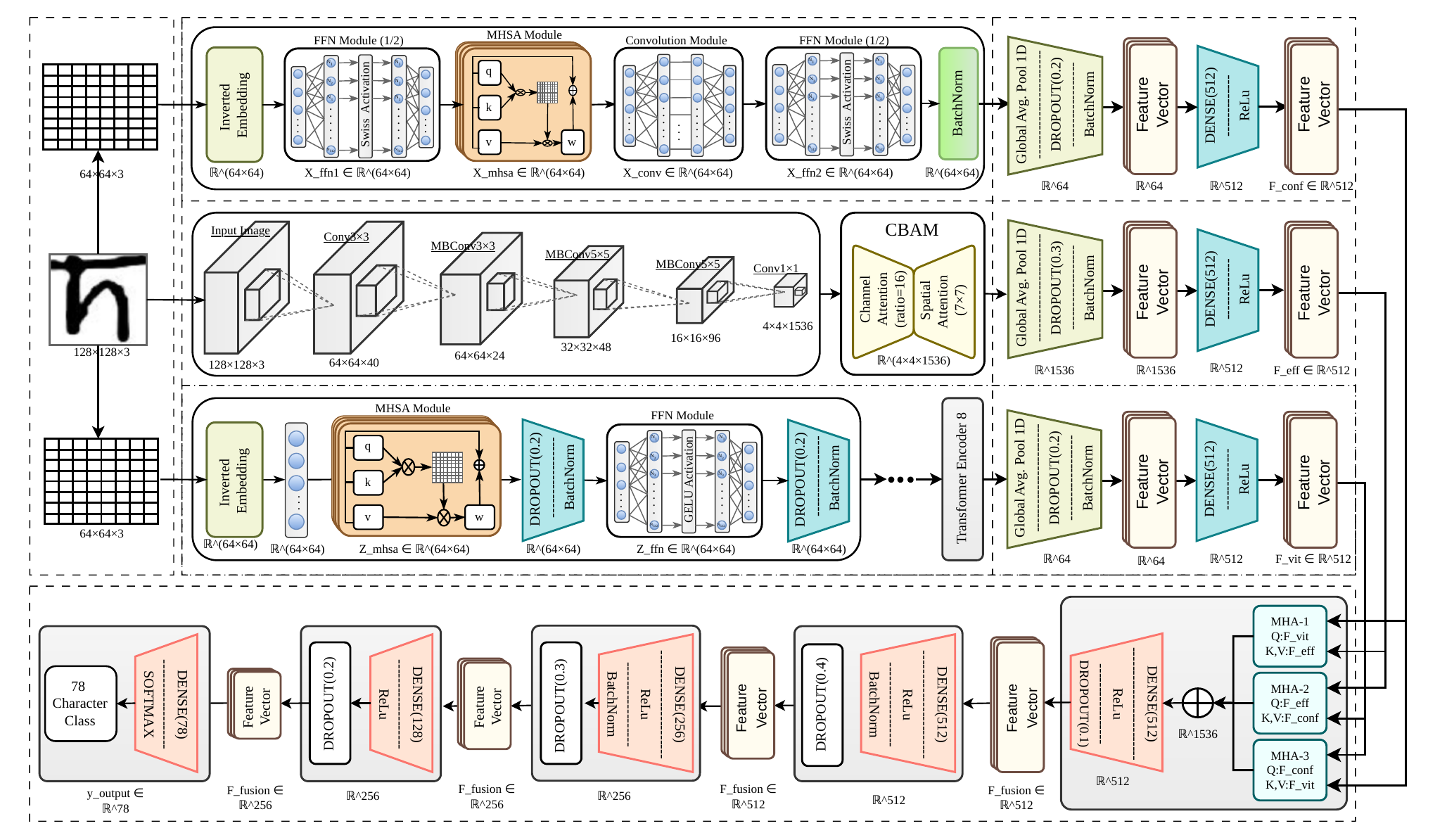}
    \caption{Model Architecture of the Proposed Bangla Handwritten Character recognition Framework}
    \label{fig-architecture}
\end{figure*}

\subsection{Proposed Methodology Overview}
We have proposed a Multi-Head Cross-Attention Fusion Network to address limitations in our earlier VashaNet versions\cite{raquib2024vashanet, raquib2024vashanet_v2} models in this study. This works This works should be focused only on basic handwritten Bangla characters. This works applied local feature extraction using CNNs, and were tested on relatively small datasets. The current work expands to 78 character classes—with numerals and compound characters—across 50,700 balanced samples. We have introduced greater intra- and inter-class complexity datasets in this research. MHCAF-Net uses a hybrid approach combining transformer-based and convolutional modelling. It leverages EfficientNetB3 for local features, Vision Transformer for long-range dependencies, and a Conformer module for multi-scale information. CBAM attention, along with a Multi-Head Cross-Attention Fusion mechanism, enhances the network’s ability to refine joint representations and specifically targets increased structural complexity. 

\subsubsection{\textbf{EfficientNetB3 + CBAM Local Feature Extraction}}
EfficientNetB3~\cite{tan2019efficientnet} serves as the convolutional backbone for local 
feature extraction. The input image is resized to $128 \times 128 \times 3$ and denoted as, \begin{equation}
X \in \mathbb{R}^{128 \times 128 \times 3},
\end{equation}
where the three dimensions represent height, width, and RGB channels, respectively. To leverage 
pretrained ImageNet representations, the first 100 layers remain frozen during training. 
EfficientNetB3 progressively reduces spatial resolution while increasing channel depth. 
The final convolutional output is \begin{equation}
F_e \in \mathbb{R}^{4 \times 4 \times 1536},
\end{equation}
where $4 \times 4$ is the reduced spatial resolution, and 1536 is the channel depth at the 
final stage. This feature tensor captures fine stroke-level structures including curves, 
intersections, and matras.

\paragraph{\textbf{MBConv Block}}
Each MBConv block expands channels, applies depthwise convolution~\cite{howard2017mobilenets}, 
and projects features back to a lower dimension. Depthwise convolution is defined as \begin{equation}
Y_{i,j,c} = \sum_{m}\sum_{n} K_{m,n,c} \cdot X_{i+m,j+n,c},
\end{equation}
where $Y_{i,j,c}$ is the output at spatial location $(i,j)$ for channel $c$, $K_{m,n,c}$ 
is the depthwise kernel for channel $c$, and $(m,n)$ indexes the kernel spatial offsets. 

\paragraph{\textbf{Convolutional Block Attention Module (CBAM)}}
CBAM~\cite{woo2018cbam} refines $F_e$ through sequential channel and spatial attention 
mechanisms, suppressing irrelevant background responses and emphasizing discriminative 
stroke regions.

\textbf{Channel Attention:} Channel attention identifies which feature channels carry 
the most relevant stroke information. Global average pooling (GAP) and global max pooling 
(GMP) are applied along the spatial dimensions to produce channel descriptors: \begin{equation}
F_{avg} = GAP(F_e), \quad F_{max} = GMP(F_e),
\end{equation} where $F_{avg}, F_{max} \in \mathbb{R}^{1 \times 1 \times 1536}$ are the spatially 
pooled descriptors. Both descriptors are passed through a shared MLP with a reduction 
ratio of 16 to compute the channel attention map\cite{woo2018cbam}:
\begin{equation}
M_c = \sigma\bigl(MLP(F_{avg}) + MLP(F_{max})\bigr),
\end{equation}
where $\sigma$ denotes the sigmoid function and $M_c \in \mathbb{R}^{1 \times 1 \times 1536}$ 
is the channel attention weight. The channel-refined feature map is obtained by 
element-wise multiplication\cite{woo2018cbam}:
\begin{equation}
F_c = M_c \otimes F_e,
\end{equation}
where $\otimes$ denotes element-wise multiplication and $F_c \in \mathbb{R}^{4 \times 4 \times 1536}$.

\textbf{Spatial Attention:} Spatial attention further refines $F_c$ by identifying 
which spatial locations contain discriminative stroke patterns. Average and max pooling 
are applied along the channel dimension to generate spatial descriptors:
\begin{equation}
F_{avg}^s = AvgPool(F_c), \quad F_{max}^s = MaxPool(F_c),
\end{equation}
where $F_{avg}^s, F_{max}^s \in \mathbb{R}^{4 \times 4 \times 1}$. The two descriptors 
are concatenated and passed through a $7 \times 7$ convolution to produce the spatial 
attention \cite{woo2018cbam}:
\begin{equation}
M_s = \sigma\bigl(f^{7\times7}([F_{avg}^s\,;\,F_{max}^s])\bigr),
\end{equation}
where $f^{7\times7}$ denotes the $7 \times 7$ convolutional layer and 
$M_s \in \mathbb{R}^{4 \times 4 \times 1}$ is the spatial attention weight. The 
spatially refined output is\cite{woo2018cbam}:

\begin{equation}
F_{out} = M_s \otimes F_c,
\end{equation}

where $F_{out} \in \mathbb{R}^{4 \times 4 \times 1536}$. Global Average Pooling and 
Dropout(0.3) are then applied to $F_{out}$ to produce the final local feature vector 
of the EfficientNetB3-CBAM branch:

\begin{equation}
F_{eff} \in \mathbb{R}^{512},
\end{equation}

where $F_{eff}$ is forwarded to the cross-attention fusion module.


\subsubsection{\textbf{Vision Transformer Based Global Feature Extraction}}

The Vision Transformer branch follows~\cite{dosovitskiy2020vit}. The image is divided 
into $16 \times 16$ patches, yielding $N = 64$ tokens. Each patch is flattened and 
linearly projected into a 64-dimensional embedding space.

\paragraph{\textbf{Patch Embedding}}

Each flattened patch $x_i \in \mathbb{R}^{p^2 \cdot C}$ is linearly projected using 
weight matrix $E \in \mathbb{R}^{(p^2 \cdot C) \times d}$ and bias $b$ to produce 
token $z_i$:

\begin{equation}
z_i = x_i E + b,
\end{equation}

where $p = 16$ is the patch size, $C = 3$ is the number of input channels, and 
$d = 64$ is the embedding dimension. Learnable positional embeddings 
$E_{pos} \in \mathbb{R}^{N \times d}$ are added to preserve spatial ordering:

\begin{equation}
Z_0 = [z_1, \dots, z_N] + E_{pos},
\end{equation}

where $Z_0 \in \mathbb{R}^{N \times d}$ is the input sequence to the first 
transformer encoder layer.

\paragraph{\textbf{Multi-Head Self-Attention}}

Query, key, and value matrices are computed from the sequence $Z$ using learned 
projection matrices $W_Q, W_K, W_V \in \mathbb{R}^{d \times d_k}$~\cite{vaswani2017attention}:

\begin{equation}
Q = ZW_Q, \quad K = ZW_K, \quad V = ZW_V.
\end{equation}

Scaled dot-product attention~\cite{vaswani2017attention} is then computed as:

\begin{equation}
Attention(Q,K,V) = Softmax\!\left(\frac{QK^T}{\sqrt{d_k}}\right)V
\label{eq:mhsa}
\end{equation}

where $d_k$ is the key dimension used for scaling to prevent vanishing gradients 
in the softmax. Multi-head attention runs $h = 4$ parallel attention heads and 
concatenates their outputs, projected by $W^O \in \mathbb{R}^{hd_v \times d}$:
\begin{equation}
MHSA(Z) = Concat(head_1, \dots, head_h)\,W^O.
\end{equation}

\paragraph{\textbf{Feed Forward Network}}
Each transformer block applies a two-layer FFN with GELU 
activation~\cite{hendrycks2016gelu}, where $W_1 \in \mathbb{R}^{d \times d_{ff}}$, 
$W_2 \in \mathbb{R}^{d_{ff} \times d}$, and inner dimension $d_{ff} = 512$:
\begin{equation}
FFN(x) = GELU(xW_1 + b_1)\,W_2 + b_2.
\end{equation}

Residual connections and LayerNorm~\cite{ba2016layernorm} stabilize learning. 
After Global Average Pooling and Dropout(0.2), the ViT branch produces the 
global feature vector:

\begin{equation}
F_{vit} \in \mathbb{R}^{512},
\end{equation}

where $F_{vit}$ encodes long-range structural dependencies across the full 
character image and is forwarded to the cross-attention fusion module.


\subsubsection{\textbf{Conformer-Based Hybrid Feature Modelling}}

The Conformer architecture~\cite{gulati2020conformer} uses self-attention and convolution in a Macaron-style structure to capture global and local dependencies. Each Conformer block proceeds as follows:

\[
\text{FFN}_{1/2} \rightarrow \text{MHSA} \rightarrow \text{Conv} \rightarrow \text{FFN}_{1/2}.
\]

Four Conformer blocks are used, each with embedding dimension $d=64$ and four attention heads.

\paragraph{\textbf{First Feed Forward Module (Macaron Half-Step)}}

The first feed-forward module uses half-step residual scaling to stabilise deep training. Let $x \in \mathbb{R}^{N \times d}$ be the input token matrix. The transformation is:

\begin{equation}
x^{(1)} 
=
x 
+
\frac{1}{2}
\Bigl(
Swish(xW_1 + b_1)W_2 + b_2
\Bigr),
\end{equation}

where $W_1 \in \mathbb{R}^{d \times d_{ff}}$, 
$W_2 \in \mathbb{R}^{d_{ff} \times d}$, 
and $d_{ff}$ denotes the feed-forward expansion dimension.

\paragraph{\textbf{Multi-Head Self-Attention}}

After the feed-forward module, Layer Normalization is applied before the attention operation (Pre-Norm). 
The Conformer block then applies multi-head self-attention (MHSA) to capture global contextual dependencies. 
The attention mechanism follows the same formulation defined in the Vision Transformer section (Eq.~\ref{eq:mhsa}.

The residual attention output is computed as:\begin{equation}
x^{(2)} = x^{(1)} + MHSA(x^{(1)}).
\end{equation}

\paragraph{\textbf{Convolution Module}}

Once attention is computed, the convolution module enhances local modeling. 
First, a pointwise convolution expands channels:

\begin{equation}
X_{gate}
=
GELU\bigl(Conv_{1\times1}(x^{(2)})\bigr),
\end{equation}

where $X_{gate} \in \mathbb{R}^{N \times 2d}$.

Depthwise convolution~\cite{howard2017mobilenets} is then applied:

\begin{equation}
Y_{i,j,c}
=
\sum_{m}\sum_{n}
K_{m,n,c}
\cdot
X_{gate,\,i+m,j+n,c}.
\end{equation}

The output is normalized and projected back:\begin{equation}
x^{(3)}
=
x^{(2)}
+
Conv_{1\times1}
\bigl(
Swish(BN(Y))
\bigr),
\end{equation}

where $x^{(3)} \in \mathbb{R}^{N \times d}$.

\paragraph{\textbf{Second Feed Forward Module (Macaron Half-Step)}}

A second half-step feed-forward module is applied:

\begin{equation}
x^{(4)}
=
x^{(3)}
+
\frac{1}{2}
\Bigl(
Swish(x^{(3)} W_3 + b_3)W_4 + b_4
\Bigr),
\end{equation}

where $W_3 \in \mathbb{R}^{d \times d_{ff}}$ and 
$W_4 \in \mathbb{R}^{d_{ff} \times d}$.

\begin{algorithm}[H]
\caption{Parallel Feature Extraction} 
\label{alg:parallel_features}
\small
\begin{algorithmic}[1]

\Require Input image $X \in \mathbb{R}^{128\times128\times3}$, Patch size $p=16$, number of patches $N=64$, Conformer blocks $L=4$

\Ensure Feature vectors $F_{eff},F_{vit},F_{conf}$

\State Initialize EfficientNetB3 backbone, CBAM module
\State Initialize Vision Transformer encoder
\State Initialize Conformer blocks

\State \textbf{Parallel Branch Execution}

\For{each branch $b \in \{\text{CNN},\text{ViT},\text{Conformer}\}$}

\If{$b=$ EfficientNet branch}

\State $F_e \gets EfficientNetB3(X)$
\State $M_c \gets \sigma(MLP(GAP(F_e)) + MLP(GMP(F_e)))$
\State $F_c \gets M_c \otimes F_e$
\State $M_s \gets \sigma(f^{7\times7}([AvgPool(F_c);MaxPool(F_c)]))$
\State $F_{eff} \gets GAP(M_s \otimes F_c)$

\EndIf

\If{$b=$ ViT branch}

\For{$i=1$ to $N$}
\State $z_i \gets x_iE + b$
\EndFor

\State $F_{vit} \gets Attention(Q,K,V)$

\EndIf

\If{$b=$ Conformer branch}

\State Initialize token sequence $x$

\For{$l=1$ to $L$}
\State $x \gets FFN_{1/2} \rightarrow MHSA \rightarrow Conv \rightarrow FFN_{1/2}$
\EndFor

\State $F_{conf} \gets x$

\EndIf

\EndFor

\State \Return $F_{eff},F_{vit},F_{conf}$

\end{algorithmic}
\end{algorithm}

\paragraph{\textbf{Layer Normalization}}

Layer normalization~\cite{ba2016layernorm} is applied after each sub-module. 
Given feature vector $x$, with mean $\mu$ and variance $\sigma^2$ computed over 
the feature dimension:

\begin{equation}
\hat{x}
=
\frac{x-\mu}{\sqrt{\sigma^2+\epsilon}},
\end{equation}

where $\epsilon$ is a stability constant.

After the four Conformer blocks, global average pooling and Dropout(0.2) produce the hybrid feature vector as,\begin{equation}
F_{conf} \in \mathbb{R}^{512}.
\end{equation} where $F_{conf}$ captures complementary local-global patterns and is forwarded to 
the cross-attention fusion module.


\subsubsection{\textbf{Multi-Head Cross-Attention Interaction Module}}
The proposed multi-head cross-attention module allows interaction between heterogeneous feature representations, unlike simple feature concatenation. Let $F_{eff} \in \mathbb{R}^{512}$, $F_{vit} \in \mathbb{R}^{512}$, and 
$F_{conf} \in \mathbb{R}^{512}$ denote the feature vectors obtained from the 
EfficientNetB3-CBAM, Vision Transformer, and Conformer branches. Each feature 
is normalized using Batch Normalization~\cite{ioffe2015batchnorm} and projected 
into a shared embedding space. Cross-attention enables information exchange between branches. One feature 
vector acts as the query while another provides the key and value. The attention computation follows the scaled dot-product formulation defined in Eq.~(\ref{eq:mhsa}). 

Three cross-attention interactions are performed to exchange information between the branches. Each interaction follows the same formulation but uses different query–key–value combinations: \begin{equation}
\begin{aligned}
MHA_i = \text{LayerNorm}\Big(&F_i + 
\text{Attention}(F_i W_Q^{(i)}, \\
&F_j W_K^{(j)}, F_j W_V^{(j)})\Big)
\end{aligned}
\end{equation}

where $(i,j)$ indicates the interacting feature pairs. The interactions are:\begin{align}
(i,j)_1 &= (vit,eff), \\
(i,j)_2 &= (eff,conf), \\
(i,j)_3 &= (conf,vit).
\end{align}
These interactions enable each branch to refine its representation using information from another branch. This process captures complementary local stroke details and global structural relationships.

The enhanced features are concatenated and fused using a fully connected 
layer with ReLU activation:

\begin{equation}
F_{fusion} =
ReLU\!\left(
W_f [MHA_1 \| MHA_2 \| MHA_3] + b_f
\right),
\end{equation}

where $W_f \in \mathbb{R}^{1536 \times 512}$ and 
$b_f \in \mathbb{R}^{512}$. The fused feature vector 
$F_{fusion} \in \mathbb{R}^{512}$ forms the unified representation 
and is passed to the classification head.

This interaction enables every branch to sharpen its representation by using complementary features and this increases the discrimination between structurally equivalent Bangla characters.


\subsubsection{\textbf{Classification Head and Training Strategy}}

The fused feature vector $F_{fusion} \in \mathbb{R}^{512}$ is passed through a
stack of fully connected layers with 512, 256, and 128 neurons, followed by a
final classification layer with 78 outputs. Each layer applies Batch
Normalization~\cite{ioffe2015batchnorm} and Dropout for regularization. The
overall transformation of the classification head can be expressed as: \begin{equation}
z = W_4 \, \phi \!\left(W_3 \, \phi \!\left(W_2 \, \phi \!\left(W_1 F_{fusion}+b_1\right)+b_2\right)+b_3\right)+b_4,
\end{equation}
where $W_1 \in \mathbb{R}^{512 \times 512}$, $W_2 \in \mathbb{R}^{512 \times 256}$,
$W_3 \in \mathbb{R}^{256 \times 128}$, $W_4 \in \mathbb{R}^{128 \times 78}$,
and $b_1,b_2,b_3,b_4$ denote the learnable bias vectors. The vector
$z \in \mathbb{R}^{78}$ represents the logits corresponding to the 78
character classes, and $\phi(\cdot)$ denotes the activation, normalization,
and dropout operations.

The final class probabilities are obtained using the Softmax
function~\cite{bridle1990softmax}

\begin{equation}
\hat{y}_i = \frac{e^{z_i}}{\sum_{j=1}^{78} e^{z_j}}.
\end{equation}

The network is trained by minimizing the categorical cross-entropy
loss~\cite{goodfellow2016deep}

\begin{equation}
\mathcal{L} = -\sum_{i=1}^{78} y_i \log(\hat{y}_i),
\end{equation}

where $y_i \in \{0,1\}$ denotes the ground-truth one-hot label and
$\hat{y}_i$ is the predicted probability.

Training uses the Adam optimizer with a learning rate of $5 \times 10^{-4}$.
ReduceLROnPlateau decreases the learning rate by a factor of $0.5$ after
seven epochs without improvement. Early stopping monitors validation accuracy
with a patience of 15 epochs. The model is trained for 40 epochs with class
weighting to address class imbalance.

\begin{algorithm}[H]
\caption{Training Procedure of the Multi-Head Cross-Attention Framework}
\label{alg:mhcaf_training}
\small
\begin{algorithmic}[1]

\Require Training images $X$, labels $Y$, Epochs $E$, batch size $B$, learning rate $\alpha$
\Ensure Training loss $\mathcal{L}$

\State Initialize network parameters $\Theta$
\State Initialize Adam optimizer

\If{training mode}

\For{epoch $=1$ to $E$}

\For{each minibatch $(X_{batch},Y_{batch})$}

\State $(F_{eff},F_{vit},F_{conf}) \gets \text{Alg.~\ref{alg:parallel_features}}(X_{batch})$

\State \textbf{Cross-Attention Fusion}

\For{$k = 1$ to $3$}
\State $MHA_k \gets \text{CrossAttention}(F_{eff},F_{vit},F_{conf})$
\EndFor

\State $F_{fusion} \gets \text{ReLU}\big(W_f \cdot 
\text{Concat}(MHA_1, MHA_2,$
\State \hspace{2.7cm}$MHA_3) + b_f\big)$

\State \textbf{Dense Classification}

\State $h_1 \gets W_1F_{fusion}+b_1$
\State $h_2 \gets W_2h_1+b_2$
\State $h_3 \gets W_3h_2+b_3$

\State $z \gets W_4h_3+b_4$

\State $\hat{y} \gets \text{Softmax}(z)$

\State \textbf{Cross-Entropy Loss}

\[
\mathcal{L} = -\sum_{i=1}^{C} y_i \log(\hat{y}_i)
\]

\State Backpropagate gradients
\State Update parameters: $\Theta \gets \Theta - \alpha \nabla_{\Theta}\mathcal{L}$

\EndFor
\EndFor

\State \Return $\mathcal{L}$

\Else

\State $\hat{y} \gets \text{Softmax}(z)$
\State Predict label: $\hat{y}_{label} = \arg\max(\hat{y})$
\State \Return $\hat{y}_{label}$

\EndIf

\end{algorithmic}
\end{algorithm}

A broad outline how features are extracted and trained will be described in Algorithm\ref{alg:parallel_features} and Agorithm\ref{alg:mhcaf_training} . It begins with a $128 \times 128 \times 3$ character image drawn by hand. There are three parallel features sought by three branches. The EfficientNetB3 in the first branch computes local stroke-level features, and CBAM attention processes them to provide more specific features, such as channel and spatial attention, to generate $F_{eff}$. The second branch is the Vision Transformer that separates the image into patches and learns the global architecture associations via self-attention to produce $F_{vit}$. The third arm applies Conformer blocks to the Feed-forward, attention and convolution operations to generate the hybrid representation $F_{conf}$. These are combined with the help of a multi-head cross-attention module. The merged representation goes through thick layers to be classified. The probabilities are calculated using softmax, and it is trained using cross-entropy loss, backpropagation, and Adam optimizer ($78$ classes).

\subsection{Performance Analysis}
The performance metrics that were used to assess the proposed architecture are accuracy, precision, recall, F1-score, and categorical cross-entropy loss. The available measures of accuracy practice the overall correctness of classification. Precision is the proportion of correctly predicted samples, and recall is the proportion of relevant samples the model identifies. Precision and recall are balanced in the F1-score. The monitors of the learning behavior were training and validation accuracy and loss curves. The curves showed gradual convergence in loss and a slow increase in accuracy. The validation loss factor helped stop early to reduce overfitting. An analysis of a confusion matrix determined the performance of each class and highlighted misrecognition with similar-stroke characters. Findings revealed equal-level accuracy and recall of the 78 classes. Furthermore, 5-fold Cross-Validation ensured that the performance was consistent across different dataset portions. Generalisation was enhanced by data augmentation. The Grad-CAM visualization showed the various significant regions the model uses for recognition.

\section{Result Analysis}
\label{sec:4}


This part illustrates the experimental analysis of the suggested framework of recognizing handwritten characters in Bangla. A number of ready-made backbone networks and hybrid architectures are tested to evaluate their capabilities in terms of handwritten character features. The experiments consist of backbone comparison, preprocessing and feature extraction analysis, ablation analysis of architectural components, and five-fold cross-validation to assess the model's stability. The standard evaluation measures shown are accuracy, precision, recall, F1-score, MCC, Kappa, error rate, and AUC. Besides this, class-by-class analysis of performance is also done to see the behavior of recognition in all the 78 categories of characters. The external Bangla handwritten data is also used to test the framework's generalization ability. Lastly, explainability analysis using Grad-CAM is provided to view the areas that contribute to the model's predictions. The subsections below present the results and discussion for each experimental setting.

\begin{table*}[ht]
\centering
\small
\resizebox{\textwidth}{!}{
\begin{tabular}{lcccccccccc}
\toprule
Model & Train (\%) & Val (\%) & Test (\%) & Error (\%) & Precision (\%) & Recall (\%) & F1 (\%) & MCC (\%) & Kappa (\%) & AUC (\%) \\
\midrule
MobileNetV2 & 98.87 & 95.94 & 96.02 & 3.98 & 96.04 & 96.02 & 96.02 & 95.98 & 95.98 & 99.96 \\
Xception & 99.02 & 96.71 & 96.80 & 3.20 & 96.81 & 96.80 & 96.80 & 96.76 & 96.76 & 99.97 \\
ConvNeXt-Small & 99.18 & 97.08 & 97.15 & 2.85 & 97.17 & 97.15 & 97.15 & 97.11 & 97.11 & 99.98 \\
Conformer & 99.12 & 96.28 & 96.36 & 3.64 & 96.35 & 96.36 & 96.35 & 96.31 & 96.31 & 99.98 \\
EfficientNet & 99.09 & 96.75 & 96.69 & 3.31 & 96.70 & 96.69 & 96.70 & 96.64 & 96.64 & 99.97 \\
ViT & 99.03 & 97.24 & 97.12 & 2.88 & 97.11 & 97.12 & 97.11 & 97.08 & 97.08 & 99.98 \\
EfficientNet + Conformer & 99.42 & 97.89 & 97.95 & 2.05 & 97.97 & 97.95 & 97.95 & 97.92 & 97.92 & 99.98 \\
ViT + EfficientNet & 99.48 & 97.96 & 98.04 & 1.96 & 98.06 & 98.04 & 98.04 & 98.01 & 98.01 & 99.98 \\
ViT + Conformer & 99.51 & 98.02 & 98.09 & 1.91 & 98.11 & 98.09 & 98.09 & 98.06 & 98.06 & 99.98 \\
EfficientNet + CBAM & 99.55 & 98.10 & 98.18 & 1.82 & 98.19 & 98.18 & 98.18 & 98.15 & 98.15 & 99.98 \\
EfficientNet + CBAM + Conformer & 99.55 & 98.10 & 98.18 & 1.82 & 98.19 & 98.18 & 98.18 & 98.15 & 98.15 & 99.98 \\
ViT + EfficientNet + CBAM & 99.63 & 98.28 & 98.35 & 1.65 & 98.37 & 98.35 & 98.35 & 98.32 & 98.32 & 99.98 \\
\rowcolor{rowblue}
\textbf{{Proposed}} & 
\textbf{{99.71}} & 
\textbf{{98.50}} & 
\textbf{{98.84}} & 
\textbf{{1.16}} & 
\textbf{{98.87}} & 
\textbf{{98.84}} & 
\textbf{{98.84}} & 
\textbf{{98.83}} & 
\textbf{{98.83}} & 
\textbf{{99.98}} \\
\bottomrule
\end{tabular}}
\caption{Performance Comparison of Backbone Models}
\label{tab:backbone_model}
\end{table*}

\subsection{Comprehensive Backbone and Fusion Evaluation}
As shown in Table \ref{tab:backbone_model}, we have studied various backbone models, including pretrained models, as well as hybrid models. MobileNetV2 and Xception achieved about 96\% test accuracy, whereas ConvNeXt-Small and ViT achieved more than 97\%. The important measures (precision, F1, MCC, Kappa) ranged from 95.9\% to 97\%. Error rates ranged from 2.85\% to 3.98\%. These findings verify that contemporary pretrained backbones can learn effective and discriminative character features. But, yet the improvement margins are inconsiderable when managed alone.

Next, we have considered the hybridization of EfficientNet with Conformer and ViT with test accuracies exceeding 97.9\%. Attention-boosted models (EfficientNet + CBAM, ViT + EfficientNet + CBAM) achieved an F1 score of 98.3984 and an error rate of 1.6-2.0\%. The MCC values were also above 98\%, indicating stronger agreement and a balanced prediction. Lastly, Our proposed did not have the highest level of variation among metrics. It had the lowest error rate of 1.16\%, and Precision, Recall, F1, MCC, and Kappa all exceeded 98.8\%. The results further show that fully hybrid fusion, along with the cross-attention, is more profoundly feature-interacting than either backbone or model stacking.

\begin{table*}[ht]
\centering
\small
\resizebox{\textwidth}{!}{
\begin{tabular}{lcccccccccc}
\toprule
Model & Train (\%) & Val (\%) & Test (\%) & Error (\%) & Precision (\%) & Recall (\%) & F1 (\%) & MCC (\%) & Kappa (\%) & AUC (\%) \\
\midrule
Proposed + Prewitt & 98.86 & 96.19 & 96.25 & 3.75 & 96.26 & 96.25 & 96.25 & 96.20 & 96.20 & 99.96 \\
Proposed + Canny & 98.98 & 96.78 & 96.82 & 3.18 & 96.83 & 96.82 & 96.82 & 96.78 & 96.78 & 99.97 \\
Proposed + Gaussian & 99.04 & 97.15 & 97.19 & 2.81 & 97.20 & 97.19 & 97.19 & 97.15 & 97.15 & 99.97 \\
Proposed + CLAHE & 99.21 & 97.56 & 97.63 & 2.37 & 97.64 & 97.63 & 97.63 & 97.59 & 97.59 & 99.98 \\
Proposed + Gaussian + Canny & 99.40 & 97.91 & 97.98 & 2.02 & 97.99 & 97.98 & 97.98 & 97.95 & 97.95 & 99.98 \\
Proposed + CLAHE + Canny & 99.56 & 98.24 & 98.30 & 1.70 & 98.31 & 98.30 & 98.30 & 98.27 & 98.27 & 99.98 \\
Proposed + Gaussian + CLAHE & 99.62 & 98.37 & 98.42 & 1.58 & 98.43 & 98.42 & 98.42 & 98.40 & 98.40 & 99.98 \\
Proposed + Canny + Prewitt & 99.65 & 98.41 & 98.47 & 1.53 & 98.48 & 98.47 & 98.47 & 98.44 & 98.44 & 99.98 \\
\rowcolor{rowblue}
\textbf{{Proposed}} & 
\textbf{{99.71}} & 
\textbf{{98.50}} & 
\textbf{{98.84}} & 
\textbf{{1.16}} & 
\textbf{{98.87}} & 
\textbf{{98.84}} & 
\textbf{{98.84}} & 
\textbf{{98.83}} & 
\textbf{{98.83}} & 
\textbf{{99.98}} \\
\bottomrule
\end{tabular}}
\caption{Performance Comparison of Preprocessing and Feature Extraction Variants}
\label{tab:preprocessing_variants}
\end{table*}

\subsection{Impact of Preprocessing and Feature Extraction Variants}

We have evaluated different preprocessing and feature extraction techniques to analyze their influence on recognition performance, as presented in Table~\ref{tab:preprocessing_variants}. Prewitt and Canny, two edge-based operators, achieved consistent test accuracies above 96\%. Performance was further enhanced beyond 97\% using intensity enhancement techniques such as CLAHE and Gaussian filtering. Additional improvements were achieved through hybrid combinations of enhancement techniques, with Gaussian + CLAHE and Canny + Prewitt achieving accuracies above 98.4\%. These findings suggest that stroke clarity and structural representation are improved by image enhancement and edge refinement.

However, the proposed model outperforms all preprocessing variants, achieving the highest test accuracy of 98.84\% and the lowest error rate. Additionally, the precision, recall, F1-score, MCC, and Kappa values are also excellent. These results demonstrate that architectural enhancement has a greater impact than preprocessing alone. The proposed framework maintains stable performance without relying heavily on aggressive preprocessing and feature extraction teachniques.

\subsection{Extended Ablation Study of Stream Design and Fusion Mechanisms}
The architecture proposed combines interaction-aware feature fusion with parallel representation learning. We have used a directed ablation study to quantitatively estimate the phenomenon of each element of the architecture. Table\ref{tab:ablation_study} summarizes the results. ConvNeXt-Small is the best on the single branch, with a 97.15\% and 2.85\% error. Dual-branch concatenation (ViT + EfficientNet) representation learning and parallel learning improve its accuracy to 98.04\% and error rate to 1.96\%, indicating the strength of parallel feature extractors.

Incorporation of attention refinement (ViT + EfficientNet + CBAM) also enhances the performance to 98.35\% accuracy and a 1.65\% error, which is a positive sign of better discrimination. The optimal result is the maximum accuracy of 98.84\% and the lowest error rate of 1.16\% which is attained by the entire interaction-driven cross-attention fusion. F1-score, Precision, Recall, MCC, and Kappa also improve and become more stable. The sequential upgrades state that parallel multi-branch learning, expressly relating to inter-branch interaction, is core in obtaining the fine-grained structure divergences of Bengali handwritten characters.

\subsection{Five-Fold Cross-Validation Analysis}

We have run a five-fold cross-validation to determine the stability and generalizability of the proposed model. Table \ref{tab:kfold_results} displays the results of the K-fold. The performance of every fold is very consistent. Accuracy lies within a close range of 98.69\% to 98.91\%. Precision, Recall, and the F1-score are also equally stable. The values of MCC and Kappa are also similar in all folds.

The mean accuracy is 98.82\% \& the variations are low across folds. In the case of macro AUC, it is always near 99.98\%. This small variance indicates that the model will be robust to varied data splits. All dataset split shows vigorous performance. These findings confirm that the model generalizes and is stable. Using the architecture, high performance is always ensured in validation environments.

\begin{table*}[ht]
\centering
\large
\resizebox{\textwidth}{!}{
\begin{tabular}{lccccccccc}
\toprule
Configuration & Streams & Accuracy (\%) & Precision (\%) & Recall (\%) & F1 (\%) & MCC (\%) & Kappa (\%) & AUC (\%) & Error (\%) \\
\midrule

\multicolumn{10}{l}{\textit{Individual Branch Baselines}} \\
\midrule
EfficientNet & Single & 96.69 & 96.70 & 96.69 & 96.70 & 96.64 & 96.64 & 99.97 & 3.31 \\
ViT & Single & 97.12 & 97.11 & 97.12 & 97.11 & 97.08 & 97.08 & 99.98 & 2.88 \\
Conformer & Single & 96.36 & 96.35 & 96.36 & 96.35 & 96.31 & 96.31 & 99.98 & 3.64 \\
ConvNeXt-Small & Single & 97.15 & 97.17 & 97.15 & 97.15 & 97.11 & 97.11 & 99.98 & 2.85 \\

\midrule
\multicolumn{10}{l}{\textit{Stream Combinations}} \\
\midrule
ViT + EfficientNet & Dual & 98.04 & 98.06 & 98.04 & 98.04 & 98.01 & 98.01 & 99.98 & 1.96 \\
ViT + EfficientNet + CBAM & Dual & 98.35 & 98.37 & 98.35 & 98.35 & 98.32 & 98.32 & 99.98 & 1.65 \\

\midrule
\multicolumn{10}{l}{\textit{Fusion Design Choice (Proposed)}} \\
\midrule
\rowcolor{rowblue}
\textbf{Proposed Model} 
& \textbf{Triple} 
& \textbf{98.84} 
& \textbf{98.87} 
& \textbf{98.84} 
& \textbf{98.84} 
& \textbf{98.83} 
& \textbf{98.83} 
& \textbf{99.98} 
& \textbf{1.16} \\
\bottomrule
\end{tabular}}
\caption{Extended Ablation Study of Stream Design and Fusion Mechanisms}
\label{tab:ablation_study}
\end{table*}

\begin{table*}[ht]
\centering
\small
\setlength{\tabcolsep}{6pt}
\renewcommand{\arraystretch}{1.1}
\resizebox{\textwidth}{!}{
\begin{tabular}{lccccccc}
\toprule
Fold & Accuracy (\%) & Precision (\%) & Recall (\%) & F1 (\%) & MCC (\%) & Kappa (\%) & AUC (\%) \\
\midrule
Fold 1 & 98.69 & 98.71 & 98.69 & 98.69 & 98.65 & 98.65 & 99.97 \\
Fold 2 & 98.76 & 98.79 & 98.76 & 98.76 & 98.73 & 98.73 & 99.98 \\
Fold 3 & 98.88 & 98.90 & 98.88 & 98.88 & 98.85 & 98.85 & 99.98 \\
Fold 4 & 98.91 & 98.93 & 98.91 & 98.91 & 98.88 & 98.88 & 99.98 \\
Fold 5 & 98.84 & 98.86 & 98.84 & 98.84 & 98.81 & 98.81 & 99.98 \\
\rowcolor{rowblue}
\textbf{Average} &
\textbf{98.82} &
\textbf{98.84} &
\textbf{98.82} &
\textbf{98.82} &
\textbf{98.78} &
\textbf{98.78} &
\textbf{99.98} \\
\bottomrule
\end{tabular}}
\caption{Five-Fold Cross-Validation Performance of the Proposed Model}
\label{tab:kfold_results}
\end{table*}

\begin{table*}[h]
\centering
\large
\resizebox{\textwidth}{!}{
\begin{tabular}{lcccccccccc}
\toprule
\textbf{Work} & \textbf{Dataset} & \textbf{Architecture} & \textbf{Dataset Size} & \textbf{Class} & \textbf{Train Acc} & \textbf{Val Acc} & \textbf{Test Acc} &  \textbf{Precision} & \textbf{Recall} & \textbf{F1} \\ \midrule

VashaNet-V1 & Primary & 26-Layer DCNN & 5,750 & 50 & 98.58 & 94.60 & 94.60 & 95.00 & 95.00 & 95.00 \\ 

VashaNet-V2 & Mixed & 19-Layer DCNN & 22,000 & 50 & 95.82 & 95.20 & 95.20 & 93.53 & 93.29 & 93.29 \\ 

\rowcolor{rowblue}
\textbf{Proposed} & \textbf{Primary} & \textbf{Hybrid Fusion} & \textbf{50,700} & \textbf{78} &  \textbf{99.71} & \textbf{98.50} & \textbf{98.84} & \textbf{98.87} & \textbf{98.84} & \textbf{98.84} \\ 
\bottomrule

\end{tabular}}
\caption{Performance Comparison of Proposed Model with Previous Works}
\label{tab:previous_work}
\end{table*}

\begin{figure}[t]
\centering
\includegraphics[
width=\linewidth,
trim=4.8cm 15cm 4cm 1.7cm,
clip
]{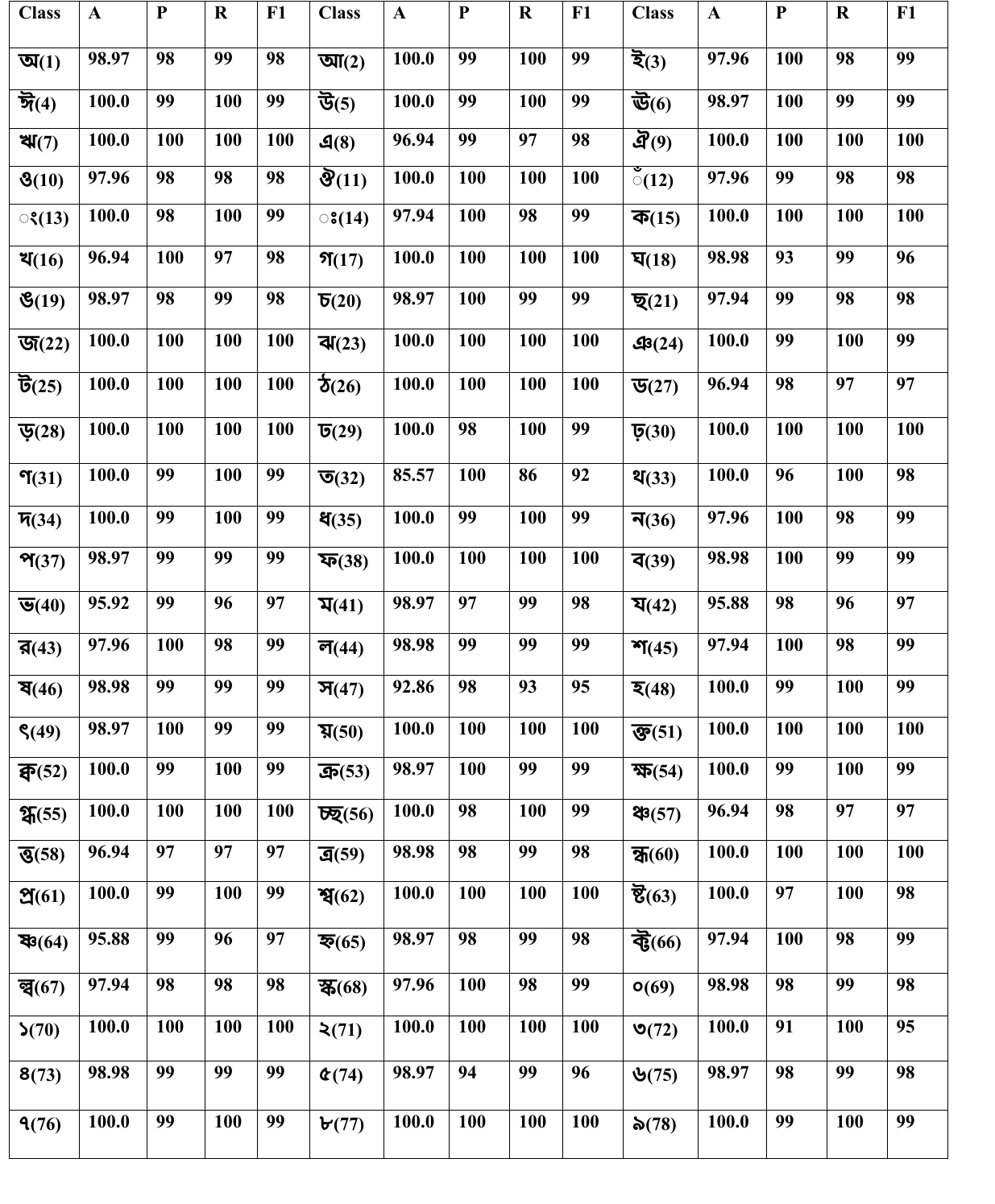}
\caption{Per-class performance results.}
\label{fig:perclass}
\end{figure}

\begin{figure}[t]
\centering
\includegraphics[
width=\linewidth]{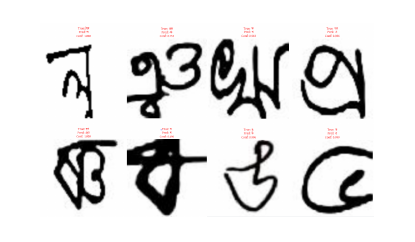}
\caption{Misclassifiaction analysis}
\label{fig:miscls}
\end{figure}

\begin{table*}[ht]
\centering
\large
\resizebox{\textwidth}{!}{
\begin{tabular}{lcccccccccc}
\toprule
Dataset & Train (\%) & Val (\%) & Test (\%) & Error (\%) & Precision (\%) & Recall (\%) & F1 (\%) & MCC (\%) & Kappa (\%) & AUC (\%) \\
\midrule
CHBCR & 99.62 & 96.65 & 96.49 & 3.51 & 96.52 & 96.49 & 96.49 & 96.45 & 96.44 & 99.91 \\
\rowcolor{rowblue}
\textbf{{Introduced}} & 
\textbf{{99.71}} & 
\textbf{{98.50}} & 
\textbf{{98.84}} & 
\textbf{{1.16}} & 
\textbf{{98.87}} & 
\textbf{{98.84}} & 
\textbf{{98.84}} & 
\textbf{{98.83}} & 
\textbf{{98.83}} & 
\textbf{{99.98}} \\
\bottomrule
\end{tabular}}
\caption{Comparision of Proposed Architecture on External CHBCR with our Introduced Dataset}
\label{tab:externaldatset}
\end{table*}

\begin{figure}[htbp]
\centering
\includegraphics[
width=\linewidth]{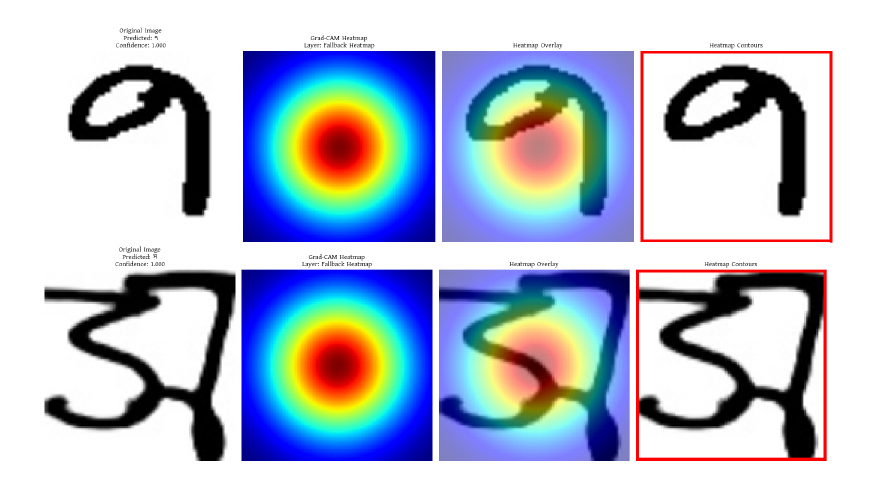}
\caption{Grad-CAM Analysis of the proposed framework.}
\label{fig:grad-cam}
\end{figure}

\subsection{Per-Class Performance and Error Analysis}

Figure~\ref{fig:perclass} represents the performance of the proposed model by classes. The model shows consistently high performance across all evaluation metrics across all classes. The number of classes with values near to 100\% shows a high recognition ability. The stroke structures with these characters are distinct and easily differentiated by the model. The findings show that the architecture works well with a large variety of character types. The model learns key structural aspects of Bangla characters, as evidenced by high scores across numerous classes. The performance table shows the detailed performance by classes.

Nonetheless, some of the characters have a few lower scores. These instances are largely found within characters who are almost too similar in script form. Bangla characters are differentiated only by a slight matra, stroke, or curve. These sections are usually depicted differently or not clearly by writers. The differences in matras, curvature, and alignment also make recognition more difficult. Figure~\ref{fig:miscls} contains several misclassified examples, which confuse the model into mistaking characters of almost the same shape. In the figure, we see that the misclassification is caused by the matras not being used properly, or, in the Juktabarno case, by any of the characters not being written properly. Such mistakes are primarily caused by the vagueness of handwriting, but not the architectural constraints. The general findings indicate that the model is stable to various styles of writing and it also performs well even when there is an elevated intra-class variance and an inter-class similarity.

\subsection{Grad-CAM based visual analysis}

Explainable Artificial Intelligence helps people understand how the model makes forecasts. Gradient-weighted Class Activation Mapping (Grad-CAM) is a method for visualizing the parts of an image that contribute to decision-making. This creates heatmaps to indicate the most crucial areas to this projected classification. Fig. grad-cam shows two predictions of the proposed model, with explanations provided by Grad-CAM. The underlinings highlight the areas the model attends to during the usualization of handwritten characters, where there is high activation around key strokes and structures. This visualization shows that the model considers stroke curves, matra position, crossroads, etc., which are discriminative among Bangla characters. Accurate predictions show a predilection for the prevailing stroke patterns. This validates that the suggested architecture does not learn from background noise but instead learns meaningful representations. Grad-CAM analysis, therefore, enhances the quality and validity of the recognition structure.

\subsection{Comparison with Previous Versions}
We compared our proposed model with our past architectures\cite{raquib2024vashanet,raquib2024vashanet_v2} and this is presented in Table\ref{tab:previous_work}. VashaNet-V1 utilised a 26-layer DCNN and came to a test accuracy of 94.60\% on the primary dataset. VashaNet-V2 was a 19-layer DCNN trained on a mixed dataset, achieving 95.20\% test accuracy. The hybrid architecture proposed performs much better on the mixed dataset than the one based on the original dataset, with a test accuracy of 98.84\%. The model does not rely entirely on deep convolutional learning. Instead, features are learned in parallel across multiple branches. It has been experimentally demonstrated that the generalization and recognition performance under large and more varied datasets is enhanced by the combination of architectural design. These findings show that the hybrid architecture is doing better compared to the past DCNN-based models. This finding indicates that the interaction-aware fusion mechanism is important in enhancing feature representation and higher recognition performance.

\subsection{Evaluation on External CHBCR Dataset}
Another evaluation that we conducted of the proposed architecture is the data on the external CHBCR\cite{towhid2020chbcrdb} dataset. The CHBCR data has handwritten Bangla vowels, consonants, numerals, and compound characters. The nature of complex stroke arrangements and structural variations produces compound characters built without the incorporation of several consonants. The CHBCR is also problematic because the Bangla script contains numerous characters that are visually similar but differ in minor movements or matras. As a result, CHBCR is not an easy dataset to character recognize. In the research, we have chosen 78 character classes out of the dataset. There are approximately 650 handwritten samples in each class, but some of the classes have slightly fewer samples. All the data consists of about 48,000 pictures. They were resized and normalized to $128\times128$. The standard train, validation and test splits were utilized to train and evaluate the model. 

Table~\ref{tab:externaldatset}, the external data set demonstrates the output on CHBCR, where there was 96.49\% test accuracy on the model. Precision, Recall and F1-score are approximately 96.5\% with MCC and Kappa of more than 96.4\%. This shows balanced and consistent performance in handwriting of different styles.

\section{Conclusion}
\label{sec:5}
This paper proposes a hybrid deep learning system for Bangla handwritten character recognition. A new balanced set of 78 Bangali character classes with 50,700 samples is presented. Contributors of various ages and socio-cultural backgrounds were used in data collection to ensure that the variations in natural writing were captured. The preprocessing pipeline enhances image quality and normalizes the dataset. The architecture suggested operates the EfficientNetB3, Vision Transformer and Conformer in parallel and the combination of their learned features with multi-head cross-attention fusion. The experiments are used to test a variety of backbone models, hybrid environments, preprocessing, and the five-fold cross-validation. The model has the highest accuracy, recalls, F1-score, MCC, and Kappa. Grad-CAM indicates important areas of stroke. Generalization is proved by external testing of the CHBCR data. The research concerns isolated characters. Further analysis of the framework will be on word and sentence recognition, multilingual handwriting analysis and lightweight real-time document processing systems.

\bibliographystyle{IEEEtran}

\end{document}